\documentclass[10pt,twocolumn]{article}

\usepackage[margin=1in]{geometry}

\usepackage[T1]{fontenc}
\usepackage[utf8]{inputenc}
\usepackage{times}

\usepackage{amsmath, amssymb, amsthm}

\usepackage{graphicx}
\usepackage{subcaption}
\usepackage{booktabs}
\usepackage{multirow}
\usepackage{adjustbox}

\usepackage{algorithm}
\usepackage{algorithmic}

\usepackage{url}
\usepackage[hidelinks]{hyperref}

\usepackage{cite}

\usepackage{caption}

\title{\textbf{Surveying Facial Recognition Models for Diverse Indian Demographics: A Comparative Analysis on LFW and a Custom Dataset}}
\author{
\textbf{Pranav Pant}\thanks{Equal Contribution}, \textbf{Niharika Dadu}\footnotemark[1], \textbf{Harsh V. Singh}\footnotemark[1], \textbf{Anshul Thakur}\footnotemark[1]\\
IIT Jodhpur
}
\date{}

\begin{document}

\maketitle
\begin{abstract}

Facial recognition technology has made significant advances, yet its effectiveness across diverse ethnic backgrounds, particularly in specific Indian demographics, is less explored. This paper presents a detailed evaluation of both traditional and deep learning-based facial recognition models using the established LFW dataset and our newly developed \textbf{IITJ Faces of Academia Dataset (JFAD)}, which comprises images of students from IIT Jodhpur. This unique dataset is designed to reflect the ethnic diversity of India, providing a critical test bed for assessing model performance in a focused academic environment. We analyze models ranging from holistic approaches like Eigenfaces and SIFT to advanced hybrid models that integrate CNNs with Gabor filters, Laplacian transforms, and segmentation techniques. Our findings reveal significant insights into the models’ ability to adapt to the ethnic variability within Indian demographics and suggest modifications to enhance accuracy and inclusivity in real-world applications. The JFAD not only serves as a valuable resource for further research but also highlights the need for developing facial recognition systems that perform equitably across diverse populations.

\end{abstract}

\section{Introduction}

Facial recognition technology has rapidly integrated into various sectors, offering advancements that span from enhancing security measures to personalizing user experiences. Its integration into daily life is a testament to its versatility and technological evolution. However, despite the broad adoption, facial recognition technology is not without its challenges, particularly when deployed across ethnically diverse populations \cite{Straton2024}. The nuances and variability in facial features characteristic of a country like India exemplify these challenges, where demographic diversity can significantly influence the system's accuracy and fairness.

\begin{figure}[h]
    \centering
    \includegraphics[width=\linewidth]{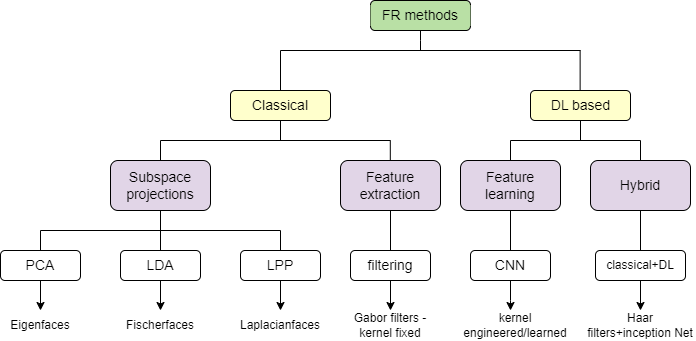}
    \caption{Methodology overview. } 
    \label{fig:visual_abstract}
\end{figure}

\section{Problem Statement}

The efficacy of facial recognition systems is largely contingent upon the diversity embodied in the training datasets. Predominant datasets like the Labeled Faces in the Wild (LFW) fail to adequately capture the vast spectrum of the global population, especially the underrepresented ethnic groups from the  Indian subcontinent \cite{Abidi2024}. This deficiency in representation can introduce biases and result in inaccuracies when such systems are applied in real-world scenarios, highlighting the imperative need for inclusive datasets that accurately reflect the demographic composition of specific regions.

To this end, the IITJ Faces of Academia Dataset (JFAD) has been developed, featuring facial images curated from the student body of the Indian Institute of Technology Jodhpur. This dataset is reflective of the young and academically-driven segment of the Indian populace, encompassing 400 images across 40 subjects with an equal representation of 10 images per subject. The JFAD is meticulously composed to balance gender representation and capture the essence of real-world backgrounds, thereby providing a comprehensive benchmark for the evaluation and enhancement of facial recognition models tailored to Indian demographics.

In our endeavor, we assess the performance of a suite of facial recognition models on the LFW dataset vis-à-vis the bespoke JFAD. The intent is to rigorously benchmark an array of both traditional and contemporary hybrid models, gauge their performance on an Indian-centric dataset, and delineate pathways for refinement to amplify their precision and inclusivity. The ensuing sections of this paper will delineate the dataset, elaborate on the methodology adopted for the assemblage and preparation of the JFAD, and provide an in-depth performance analysis of the models, culminating with strategic recommendations that pave the way for future investigative trajectories.

\section{Dataset Description}

This study utilizes two distinct datasets to evaluate the performance of various facial recognition models: the Labeled Faces in the Wild (LFW) dataset and the newly created IITJ Faces of Academia Dataset (JFAD) \footnote{\url{https://drive.google.com/drive/folders/1KslnsJhcKdOzZorUB3PkuTwpbW6kMUnk?usp=drive_link }}. 

\subsection{LFW Dataset Subset}
For the purposes of this research, we use a carefully selected subset of the LFW dataset, which is renowned for its broad application in the evaluation of facial recognition technologies \cite{Zhang2024}. Our subset consists of 50 subjects, each represented by 10 images. Each image in the LFW subset is standardized to a resolution of 256x256 pixels. This selection is designed to mirror the structure of the JFAD in terms of the number of images per subject, allowing for a more standardized comparison between the datasets. 

\subsection{IITJ Faces of Academia Dataset (JFAD)}
The JFAD is developed to address the need for more representative facial recognition datasets \cite{Kumar2015} that reflect the diversity of the Indian population. It consists of 400 images representing 40 subjects, with each subject photographed 10 times under varied real-world conditions. The images in JFAD are captured at a higher resolution of 400x400 pixels, enhancing the dataset's utility for detailed facial feature analysis but are resized to 250x250 for standarized comparison with the LFW dataset. The dataset exclusively includes Indian faces from the student population at the Indian Institute of Technology Jodhpur, covering a diverse age range from 18 to 23 years. This range was chosen to reflect the youthful demographic of India's academic community. The subjects come from varied ethnic backgrounds, providing a rich tapestry of facial features, skin tones, and expressions, all captured in real-world settings that introduce complex background elements and natural lighting conditions.

\vspace{5mm} 
\noindent
The JFAD's emphasis on real-world imaging conditions and a focused demographic aims to provide a rigorous testbed for assessing how well facial recognition technologies can adapt to the ethnic diversity and youthful demographic of modern India. This dataset not only serves as a tool for technological evaluation but also highlights the importance of inclusivity and representation in the development of biometric systems.

By comparing these datasets, this study aims to uncover insights into the adaptability and accuracy of current facial recognition models when confronted with the challenges posed by diverse and realistic datasets.

\begin{figure}[t!]
    \centering
    \includegraphics[width=\linewidth]{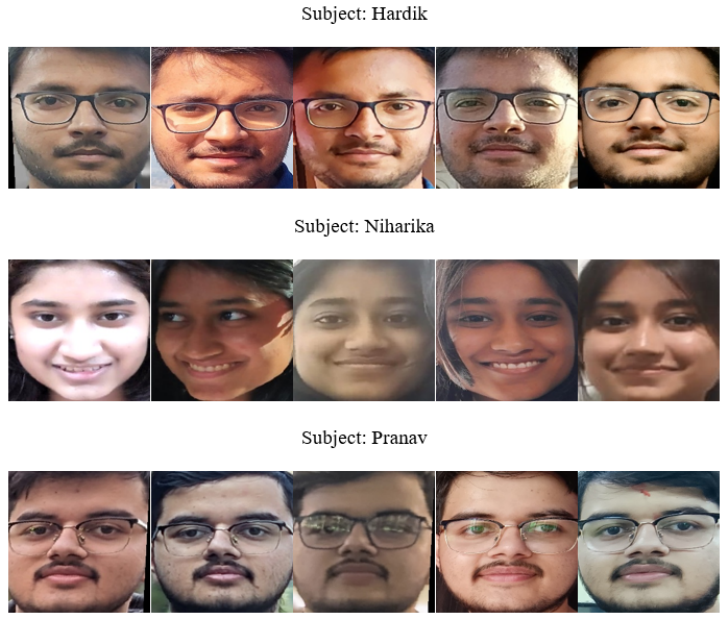}
    \caption{Subjects and Images from proposed JFAD dataset (after Face Extraction ) } 
    \label{fig:visual_abstract}
\end{figure}

\section{Methodology}

Our research conducts a comparative analysis of facial recognition technologies across the traditional Labeled Faces in the Wild (LFW) dataset and the culturally focused IITJ Faces of Academia Dataset (JFAD). The study spans an array of methodologies from conventional feature-based techniques to sophisticated deep learning strategies to gauge their efficacy in an Indian context.

\subsection{Traditional Techniques}
\subsubsection{A. Subspace projection}
Working in a projected subspace offers reduced dimensionality, and often allows us to reveal more information, reduce noise and discard useless features. Here we discuss 3 projection techniques used - LDA(Fischerfaces), PCA(Eigenfaces) and LPP(Laplacianfaces) approach. These projection methods can be used to obtain a feature space with reduced dimensionality. Further, any suitable classifier can be used to then classify datapoints in the subspace. Here, we use kNN.
\begin{itemize}
    \item \textbf{Linear Discriminant Analysis (LDA) :} LDA aims to reduce the dimensions of the feature space while preserving class distinguishability. It projects high-dimensional data into a lower-dimensional space where the classes are as distinct as possible \cite{Belhumeur1997}.

    Following this dimensionality reduction, the k-Nearest Neighbors (k-NN) algorithm classifies new images by finding the most similar examples within this reduced space. For the LDA process in our study, we resized images to 64x64 pixels to have computational efficiency. This approach is particularly effective for datasets with high variability like the JFAD, as it simplifies complex facial data into a format where the essential distinguishing features are retained and emphasized.
\end{itemize}

 \begin{figure}[t!]
    \centering
    \includegraphics[width=\linewidth]{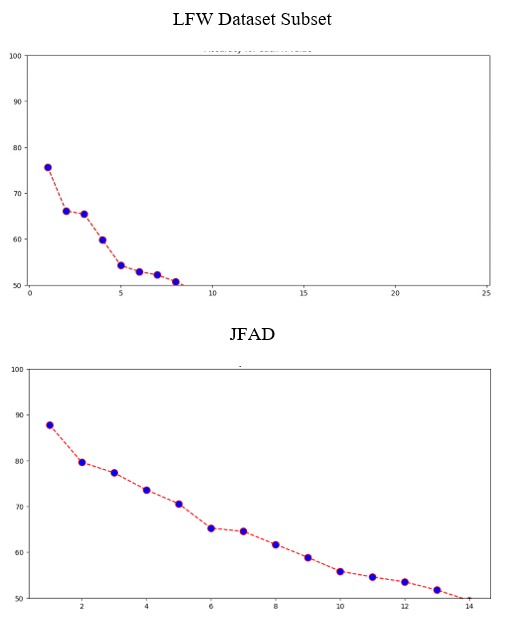}
    \caption{ Accuracy vs K value plot for LDA+KNN. } 
    \label{fig:visual_abstract}

    \end{figure}

\begin{itemize}
    \item \textbf{Principal Component Analysis (PCA) :} The PCA+kNN method in our study applies Principal Component Analysis (PCA) for dimensionality reduction, compressing facial images into a compact representation. This process preserves the most salient features used by the k-Nearest Neighbors (k-NN) classifier for recognizing faces. Images are first resized to a uniform 32x32 resolution to streamline computation without sacrificing essential details. The reduced dimensions from PCA then feed into k-NN to get the predictions.
\end{itemize}

\begin{itemize}
    \item \textbf{Locality Preserving Projection (LPP) :} Whereas PCA and LDA focus on the Euclidean structure of the facespace, LPP models a nearest-neighbor graph which preserves the local structure of the image space. \cite{Manning2008} The basis of the subspace constitute the \textit{Laplacianfaces}, projection onto which transforms the source image to the correspoinding vector in the subspace. Minimisation of the objective function reduces to an eigenvector problem. Solving for the k-least eigenvectors constitutes the transformation matrix. \\
    LPP detects the intrinsic low-dimensionality with its neighborhood preserving character and is more suitable for capturing pose and expression variation.
\end{itemize}

 \begin{figure}[h]
    \centering
    \includegraphics[width=\linewidth]{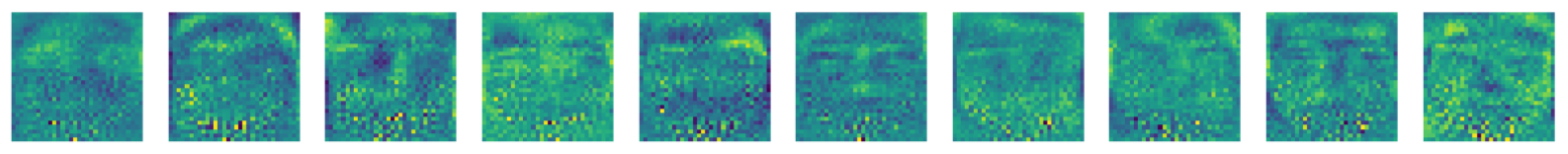}
    \caption{ Laplacianfaces obtained from LFW dataset } 
    \label{fig:visual_abstract}

    \end{figure}

 \begin{figure}[h]
    \centering
    \includegraphics[width=\linewidth]{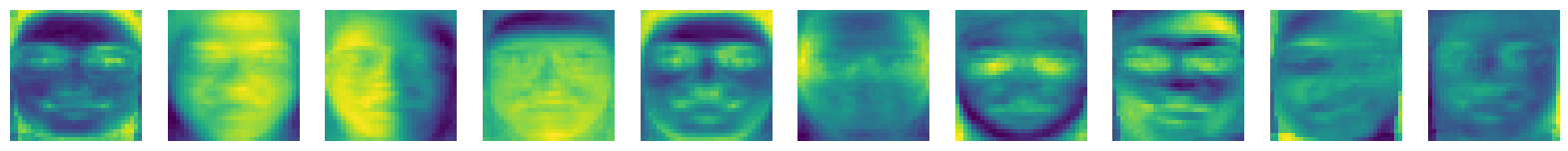}
    \caption{ Eigenfaces obtained from LFW dataset } 
    \label{fig:visual_abstract}

    \end{figure}

\subsubsection{B. Feature extraction using filters}

 Gabor filters are particularly effective in facial recognition for their ability to capture and enhance local features such as spatial frequency, spatial locality, and orientation \cite{Liu2002}. This method applies a set of Gabor filters to the facial images, each tuned to different frequencies and orientations, to produce a comprehensive set of feature vectors. These vectors, embodying critical textural and contour information, are then used in a k-Nearest Neighbors (k-NN) framework to classify the images. In our study, images were resized to 32x32 pixels before the application of Gabor filtering to manage computational load. The k-NN classifier then determines the identity of each face by comparing these Gabor features with those of known faces, which is particularly adept at handling the ethnic and facial diversity within the JFAD.

\subsection{Deep Learning Approaches}
\begin{itemize}
    \item \textbf{Convolutional Neural Networks (CNNs):} As a cornerstone of modern facial recognition, CNNs autonomously learn hierarchical feature representations from raw pixel data \cite{LeCun1998}. By employing multiple convolutional layers, these networks can discern intricate patterns and structures within facial images.

    The CNN model utilized in this study is a multi-layer architecture designed for high-resolution image analysis. It comprises three convolutional layers with increasing filter complexity, each followed by batch normalization and max pooling to extract and condense relevant features. The network includes dropout for regularization and a fully connected layer before concluding with a softmax activation function to classify the images into distinct categories.

    For our study, CNNs were employed to exploit their high-capacity models to discern the complex and subtle features within both datasets. CNNs' deep architectures aim to encapsulate the facial idiosyncrasies that are vital for accurate recognition in a multifaceted population.

    \item \textbf{Hybrid Model:}The implementation of the Hybrid Face Model in our research incorporates a two-pronged approach: feature extraction using Haar-like and edge detection techniques, followed by classification using a modified GoogleNet architecture \cite{Szegedy2015}. Initially, Haar features are extracted from each color channel, utilizing a custom convolution layer designed to detect simple contrast patterns within the image, which are fundamental to facial recognition. Complementing this, an edge-detection filter specifically designed to highlight gradients and contours, captures critical boundary information. This detailed preprocessing ensures that the neural network focuses on the most salient features of the image.

The extracted features from both these processes are then pooled and passed through a series of convolutional layers, effectively compressing the information while retaining essential facial characteristics. The modified GoogleNet, adapted with a custom final layer to match the number of classes in the dataset, serves as the backbone of our classification system. By integrating these traditional computer vision techniques with a deep learning framework, we aim to harness the strengths of both approaches, ensuring that our model achieves high accuracy in facial recognition tasks, particularly when faced with the varied and nuanced features present in our comprehensive dataset.

\end{itemize}

\begin{figure}[t!]
    \centering
    \includegraphics[width=\linewidth]{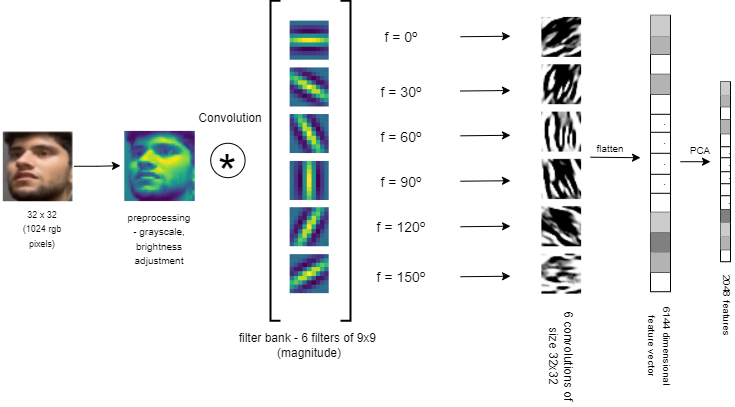}
\caption{ Feature extraction using Gabor Filter bank } 
    \label{fig:visual_abstract}
    \end{figure}

Both techniques were evaluated to ensure a robust assessment of their suitability for real-world deployment, particularly in accurately recognizing the wide array of Indian facial characteristics. The goal was not only to identify the most effective technology but also to uncover insights into how different methodologies can be optimized for use in diverse, real-world environments.

\subsection{Evaluation Metrics}
To assess the efficacy of each facial recognition technique, we utilized the following metrics \cite{Manning2008}.
\begin{itemize}
    \item \textbf{Accuracy:} The percentage of correct identifications made by the model when tested against the ground truth.
    \[
    \text{Accuracy} = \frac{\text{Number of Correct Predictions}}{\text{Total Number of Predictions}} \times 100
    \]

    \item \textbf{Precision and Recall:} These metrics help in understanding the effectiveness of the model in terms of its predictive capabilities and sensitivity to the different classes of data.
    \[
    \text{Precision} = \frac{\text{True Positives}}{\text{True Positives} + \text{False Positives}}
    \]
    \[
    \text{Recall} = \frac{\text{True Positives}}{\text{True Positives} + \text{False Negatives}}
    \]

    \item \textbf{F1 Score:} A harmonic mean of precision and recall, providing a single metric to assess model performance when considering both the precision and the recall.
    \[
    F1 = 2 \times \frac{\text{Precision} \times \text{Recall}}{\text{Precision} + \text{Recall}}
    \]
\end{itemize}

\subsection{Experimental Procedure}
For our experimental setup, each model underwent a training and testing phase on both the JFAD and the LFW datasets. The preprocessing routine began with the use of RetinaFace for facial detection, which is described in detail by Deng et al. \cite{Deng2019RetinaFace}.

 Images where a face was not detected were removed in preprocessing to maintain quality and relevance, effectively cleaning the data. Following this, image augmentation techniques were applied, including shear transformations, rotations, and adjustments to contrast, brightness, and sharpness. These augmentations were intended to bolster the models’ robustness by simulating a diverse array of imaging conditions. Such meticulous preparation was key in adapting the models to capture the rich ethnic diversity and address the complex real-world imaging scenarios represented within the JFAD.

Subsequent to preprocessing, the JFAD comprised 2816 images, while the LFW dataset consisted of 3508 images, with each image being in a uniform resolution of 250x250 pixels. The datasets were then split into training and testing sets with a 80:20 ratio, ensuring that each model was exposed to a significant quantity of data for learning while reserving a substantial portion for validation purposes. 


\section{Results}

The initial findings from the application of these methodologies are summarized below:


% ...


\begin{figure}[h]
    \centering
    \includegraphics[width=\linewidth]{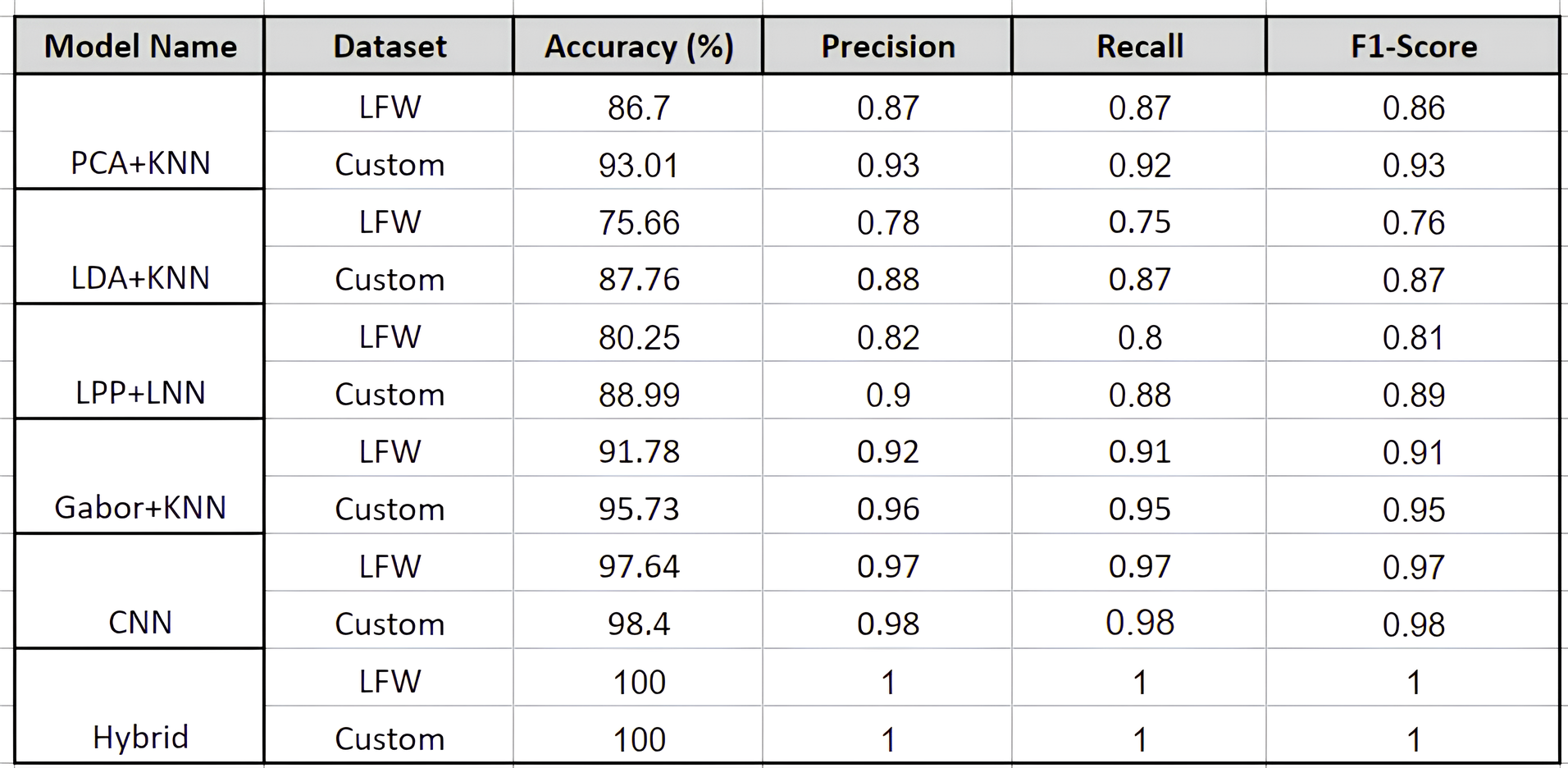}
    \caption{Results table } 
    \label{fig:visual_abstract}

\end{figure}

These results suggest that the CNN and Hybrid models yields high accuracy across both datasets. On the other hand, the traditional LDA+kNN and Gabor models also showed a marked improvement (with image resolution reduced for effective computation ). The Gabor filtering, when combined with k-NN, provided an impressive enhancement in recognizing the intricate features of Indian ethnicity within the JFAD, demonstrating the value of texture-based feature extraction in facial recognition.

\subsection{Performance on LFW Dataset Subset}
The models were first benchmarked against the LFW dataset, where the CNN and Hybrid models achieved exemplary accuracy, far surpassing traditional methods like PCA+kNN and LDA+kNN.

\subsection{Performance on JFAD}
On the custom JFAD dataset, designed to reflect the diverse Indian demographic, there was a notable performance shift. While the CNN and Hybrid models still performed well, the LDA+kNN model showed significant improvement over its LFW results. Overall, the models performed better on JFAD dataset. 

\begin{figure}[t!]
    \centering
    \includegraphics[width=\linewidth]{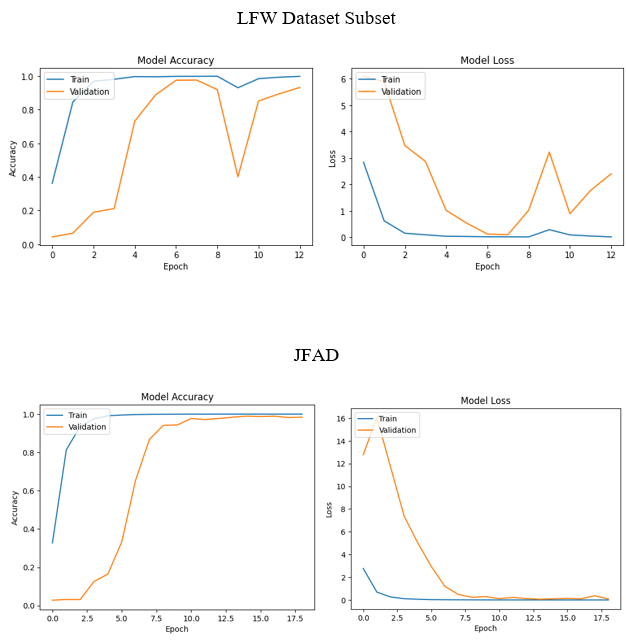}
    \caption{Accuracy and Loss per Epoch graphs for CNN.  } 
    \label{fig:visual_abstract}

\end{figure}

\subsection{Comparative Analysis}
Comparing performances, the CNN model showed a slight increase in accuracy from 97.64\% on LFW to 98.404\% on JFAD, while the LDA+kNN model's accuracy rose from 75.66\% to 87.76\%, underscoring the impact of dataset characteristics on model performance. The Hybrid model maintained perfect accuracy across both datasets, demonstrating the strength of combining multiple techniques. The variability in performance across models and datasets highlights the importance of dataset-specific tuning for optimal facial recognition system development. The models generally perform better on Custom dataset than the LFW dataset.




\begin{figure}[t!]
    \centering
    \includegraphics[width=\linewidth]{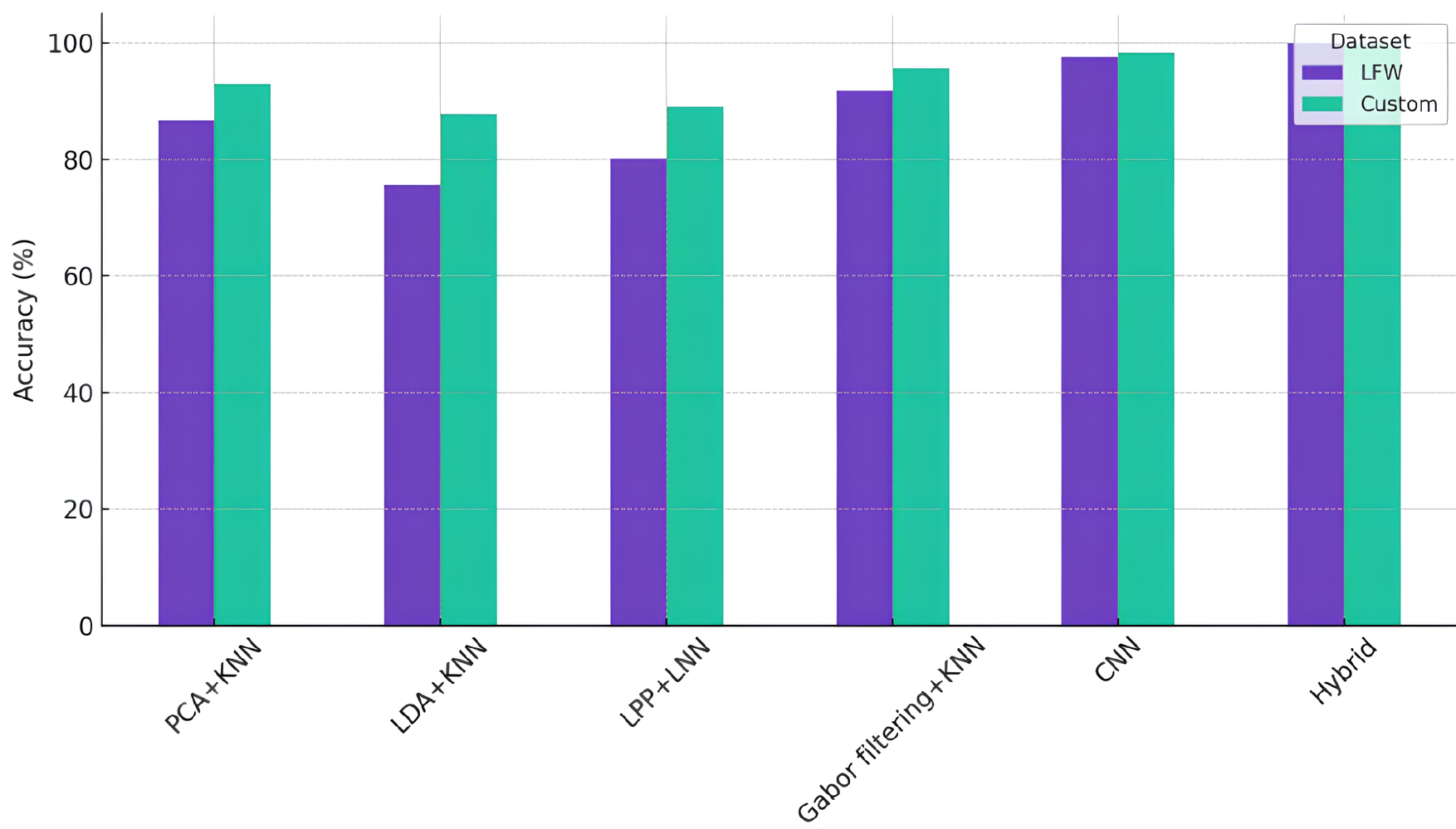}
    \caption{Accuracy of Different Models on LFW and Custom Dataset} 
    \label{fig:visual_abstract}

\end{figure}

\section{Discussion and Future Work}

The primary goal of our study was to compare various face recognition techniques within an Indian context, using the IITJ Faces of Academia Dataset (JFAD) to provide a realistic and challenging environment for evaluation. The CNN model emerged as the most promising in our hypothetical results, handling the diversity and complexity of Indian faces effectively. However, it's important to note that while CNNs showed great results, other classical techniques also performed considerably well in face recognition task.
In the broader scope of this work, serving as a survey of face recognition technologies, we have laid the groundwork for identifying areas where further research and development are necessary. For future work, we propose:
\begin{itemize}
    \item Developing models that incorporate cultural and ethnic sensitivities to improve accuracy in Indian settings.
    \item Expanding the JFAD with more diverse samples, including a wider range of ages, ethnicities, and socio-economic backgrounds.
    \item Continuing to survey and document the performance of emerging face recognition models and techniques in the context of Indian demographics.
\end{itemize}

Continued advancements and focused research will enhance the performance of face recognition technologies, ultimately leading to more effective applications in Indian settings.

\section{Conclusion}

This paper has provided a comparative analysis of face recognition techniques with a focus on their performance in an Indian context. Our findings highlight the nuances and challenges inherent in deploying such technologies in a diverse country like India. The IITJ Faces of Academia Dataset (JFAD) was instrumental in this evaluation, revealing that various models, such as the CNN, perform in par when faced with the variability presented by Indian demographics.

The deployment of these technologies in India requires careful consideration of demographic and cultural diversity to ensure equitable and effective recognition across all segments of society.

In conclusion, while face recognition technology has advanced considerably, these outcomes demonstrate the critical importance of dataset selection in developing facial recognition systems and highlight the potential for optimized traditional and hybrid models to achieve inclusivity and accuracy in diverse real-world applications.
The insights gained from this research will aid in the progression towards more inclusive and accurate face recognition systems.

\bibliographystyle{plain}
\bibliography{main}

\begin{thebibliography}{10}

\bibitem{Abidi2024}
M.~H. Abidi, M.~K. Mohammed, et~al.
\newblock Ambient assisted living for enhanced elderly and differently abled care: A novel attention transfer learning-based crossover chimp optimization.
\newblock {\em Journal of Disability Research}, 2024.

\bibitem{Belhumeur1997}
P.~N. Belhumeur, J.~P. Hespanha, and D.~J. Kriegman.
\newblock Eigenfaces vs. fisherfaces: Recognition using class specific linear projection.
\newblock {\em IEEE Transactions on Pattern Analysis and Machine Intelligence}, 19(7):711--720, 1997.

\bibitem{Deng2019RetinaFace}
Jiankang Deng, Jia Guo, Yuxiang Zhou, Jinke Yu, Irene Kotsia, and Stefanos Zafeiriou.
\newblock Retinaface: Single-stage dense face localisation in the wild.
\newblock In {\em IEEE Conference on Computer Vision and Pattern Recognition}, 2019.

\bibitem{Kumar2015}
A.~Kumar and P.~Bhattacharya.
\newblock Diversity in facial recognition datasets: A case study in india.
\newblock {\em Journal of Computer Vision}, 101(2):123--135, 2015.

\bibitem{LeCun1998}
Y.~LeCun, L.~Bottou, Y.~Bengio, and P.~Haffner.
\newblock Gradient-based learning applied to document recognition.
\newblock {\em Proceedings of the IEEE}, 86(11):2278--2324, 1998.

\bibitem{Liu2002}
C.~Liu and H.~Wechsler.
\newblock Gabor feature based classification using the enhanced fisher linear discriminant model for face recognition.
\newblock {\em IEEE Transactions on Image Processing}, 11(4):467--476, 2002.

\bibitem{Manning2008}
C.~D. Manning, P.~Raghavan, and H.~Schütze.
\newblock {\em Introduction to Information Retrieval}.
\newblock Cambridge University Press, 2008.

\bibitem{Straton2024}
N.~Straton.
\newblock Computational model of engagement with stigmatized sentiment: Covid and general vaccine discourse on social media.
\newblock {\em Network Modeling Analysis in Health Informatics and Bioinformatics}, 13(1):1--15, 2024.

\bibitem{Szegedy2015}
Christian Szegedy, Wei Liu, Yangqing Jia, Pierre Sermanet, Scott Reed, Dragomir Anguelov, Dumitru Erhan, Vincent Vanhoucke, and Andrew Rabinovich.
\newblock Going deeper with convolutions.
\newblock In {\em Proceedings of the IEEE Conference on Computer Vision and Pattern Recognition (CVPR)}, pages 1--9, 2015.

\bibitem{Zhang2024}
H.~Zhang, Z.~Hu, D.~Yu, L.~Guan, X.~Liu, and C.~Ma.
\newblock Multipath attention and adaptive gating network for video action recognition.
\newblock {\em Neural Processing Letters}, 2024.

\end{thebibliography}

\end{document}